\documentclass[conference]{IEEEtran}
\IEEEoverridecommandlockouts

\usepackage{cite}
\usepackage{amsmath,amssymb,amsfonts}
\usepackage{graphicx}
\usepackage{textcomp}
\usepackage{xcolor}
\usepackage{booktabs}
\usepackage{multirow}
\usepackage{array}
\usepackage{url}
\usepackage{comment}
\usepackage[T1]{fontenc}
\usepackage[utf8]{inputenc}
\usepackage[hidelinks]{hyperref}
\usepackage[acronym]{glossaries}
\glsdisablehyper
\usepackage{siunitx}
\DeclareSIUnit{\pp}{pp}

\def\BibTeX{{\rm B\kern-.05em{\sc i\kern-.025em b}\kern-.08em
    T\kern-.1667em\lower.7ex\hbox{E}\kern-.125emX}}

\makenoidxglossaries
\newacronym{bsn}{BSN}{Body Sensor Networks}
\newacronym{imu}{IMU}{inertial measurement unit}
\newacronym{har}{HAR}{human activity recognition}
\newacronym{loso}{LOSO}{leave-one-subject-out}
\newacronym{wmsd}{WMSD}{work-related musculoskeletal disorder}
\newacronym{ml}{ML}{machine learning}
\newacronym{gan}{GAN}{generative adversarial network}
\newacronym{ctgan}{CTGAN}{Conditional Tabular GAN}
\newacronym{lstm}{LSTM}{long short-term memory}
\newacronym{cnn}{CNN}{convolutional neural network}
\newacronym{rf}{RF}{Random Forest}
\newacronym{et}{ET}{Extra Trees}
\newacronym{xgb}{XGB}{XGBoost}
\newacronym{lda}{LDA}{Linear Discriminant Analysis}
\newacronym{lr}{LR}{Logistic Regression}
\newacronym{svm}{SVM}{Support Vector Machine}
\newacronym{pca}{PCA}{Principal Component Analysis}

\begin{document}

% Generated numerical macros (paper-pipeline reproducibility convention):
% every number below is produced by scripts/generate_all.py from
% results/revision/*.csv -- do not hand-edit build/values.tex.
\newcommand{\PerSubjMin}{61.9}
\newcommand{\PerSubjMax}{88.7}
\newcommand{\PerSubjMean}{77.5}
\newcommand{\PerSubjStd}{8.3}
\newcommand{\PerSubjSpread}{27}
\newcommand{\MainClassifier}{SVM}
\newcommand{\MainMacroFScore}{75.1}
\newcommand{\WeakUserList}{8, 9, 3}
\newcommand{\StrongUserList}{10, 13, 6}
\newcommand{\WeakAccMin}{61.9}
\newcommand{\WeakAccMax}{70.5}
\newcommand{\StrongAccMin}{86.2}
\newcommand{\StrongAccMax}{88.7}
\newcommand{\WinTwoAcc}{68.8}
\newcommand{\WinFiveAcc}{77.5}
\newcommand{\WinTenAcc}{82.2}
\newcommand{\WinGainPP}{13.4}
\newcommand{\BestSingleSensor}{Back}
\newcommand{\AllVsBestSingleGainPP}{9.6}
\newcommand{\TrouserAccDropPP}{11.3}
\newcommand{\TopConfusedPairA}{Walking$\rightarrow$Climbing}
\newcommand{\TopConfusedPairB}{HandsUp$\rightarrow$Painting}
\newcommand{\MaxHparamDelta}{1.1}
\newcommand{\StdAugDeltaMinSub}{-0.1}
\newcommand{\StdAugDeltaMaxSub}{0.7}
\newcommand{\StdAugDeltaMinFull}{-0.1}
\newcommand{\StdAugDeltaMaxFull}{0.4}
\newcommand{\CtganDeltaMin}{-5.5}
\newcommand{\CtganDeltaMax}{-0.1}
\newcommand{\CtganWorstClf}{LDA}
\newcommand{\CtganKSSigPct}{100}
\newcommand{\CtganKSSigCount}{294}
\newcommand{\CtganKSTotalCount}{294}
\newcommand{\CnnMeanAcc}{55.7}
\newcommand{\CnnMeanMacroFScore}{51.9}
\newcommand{\PersWeakAccNone}{68.4}
\newcommand{\PersWeakAccThirty}{85.2}
\newcommand{\PersWeakGainPP}{16.8}

\title{Towards Field-Deployable Wearable IMU Systems for
       Occupational Activity Recognition
}

\author{%
  \IEEEauthorblockN{Ihsan Ul Haq\textsuperscript{*}}
  \IEEEauthorblockA{%
    \textit{Centre for Computational Science}\\
    \textit{and Mathematical Modelling}\\
    Coventry University, UK\\
    haqi8@uni.coventry.ac.uk}
  \and
  \IEEEauthorblockN{James Brusey}
  \IEEEauthorblockA{%
    \textit{Centre for Computational Science}\\
    \textit{and Mathematical Modelling}\\
    Coventry University, UK\\
    aa3172@coventry.ac.uk}
  \and
  \IEEEauthorblockN{Farnad Nasirzadeh}
  \IEEEauthorblockA{%
    \textit{School of Architecture and}\\
    \textit{Built Environment}\\
    Deakin University, Australia\\
    farnad.nasirzadeh@deakin.edu.au}
  \and
  \IEEEauthorblockN{Chandan Karmakar}
  \IEEEauthorblockA{%
    \textit{School of Information Technology}\\
    Deakin University, Australia\\
    karmakar@deakin.edu.au}
}

\maketitle

% ============================================================
\begin{abstract}
Wearable \gls{imu} systems for
construction worker \gls{har} routinely
report aggregate accuracies above \SI{95}{\percent}, rarely under
genuine subject-independent (\gls{loso}) evaluation, masking the
per-individual variance that determines real-world deployment
performance.
Using the VTT-ConIoT benchmark, we conduct,
to our knowledge, the first systematic configuration sensitivity
analysis of this dataset: six classifiers across five sensor
placements and three window sizes under a single \gls{loso}
protocol, a
personalised-calibration pilot, and a validated comparison of
tabular \acrshort{gan}-based and standard time-series
augmentation. Per-subject \acrshort{loso} accuracy for the main
multi-sensor, 16-class configuration ranges from
\SIrange{\PerSubjMin}{\PerSubjMax}{\percent} (mean \SI{\PerSubjMean}{\percent}), a
\SI{\PerSubjSpread}{\pp} spread, though all data are
laboratory-collected.
Multi-sensor fusion and longer windows both improve accuracy;
standard augmentation gives small, consistent gains, while
\gls{ctgan} augmentation actively \emph{harms} accuracy with a
demonstrated distributional-fidelity mechanism; and personalised
calibration recovers \SI{\PersWeakGainPP}{\pp} for the weakest
participants from just \SI{30}{\percent} of their own data. These findings
provide evidence-grounded configuration guidance for scaling
occupational body sensor networks towards field deployment.
\end{abstract}

\begin{IEEEkeywords}
body sensor networks; \acrshort{imu}; \acrshort{har};
construction workers; \acrshort{loso};
sensor placement; configuration sensitivity; synthetic data
augmentation; occupational health; personalised monitoring
\end{IEEEkeywords}

% ============================================================
\section{Introduction}
\label{sec:intro}

\Glspl{wmsd} remain a leading cause of long-term occupational
disability in construction, driven by repetitive lifting,
awkward postures, and prolonged whole-body vibration
\cite{sensors2025ergonomics}. Wearable \glspl{imu}, combined with
supervised \gls{ml}, offer a scalable, low-cost pathway to
continuous in-field ergonomic monitoring. Such systems can classify
worker activity with sufficient fidelity to trigger real-time alerts
and safety coaching. This is a key step towards the personalised
health ecosystems envisioned by the \gls{bsn} community.

The VTT-ConIoT dataset \cite{makela2022vtt} is one of the few
publicly available multi-sensor \acrshort{imu} benchmarks designed
specifically for construction worker \gls{har}, which has attracted five
independent deep-learning studies since its 2022
release~\cite{mekruksavanich2022lstm,mekruksavanich2023deep,%
mekruksavanich2025resnet,lamsa2022video2imu,chen2025geometry},
each reporting accuracy up to $\approx$\SI{99}{\percent} and suggesting that
the classification problem is nearly solved.
We argue that this conclusion is premature, for two reasons
detailed in Section~\ref*{sec:related}: most of these figures are
not obtained under genuine subject-independent evaluation, and
none report per-subject variability.
Three critical questions remain for this benchmark, and for
construction \acrshort{imu} systems more broadly:
{\setlength{\topsep}{0pt}\setlength{\partopsep}{0pt}
\begin{enumerate}
  \item Are high aggregate accuracies stable across \emph{individual} workers, or do they conceal large per-subject variance?
  \item How do sensor placement, window length, and classifier choice jointly determine field-relevant performance?
  \item Can synthetic data augmentation reduce the per-subject gap for practitioners who cannot collect large cohorts?
\end{enumerate}}

Addressing these questions means translating body sensor
networks \emph{from lab to life}: a system with high aggregate
accuracy in a controlled setting may still perform unreliably
for specific individuals in the field when that aggregate is not
subject-independent. This paper makes three contributions:
{\setlength{\topsep}{0pt}\setlength{\partopsep}{0pt}
\begin{enumerate}
\item To our knowledge, the first per-subject \acrshort{loso} decomposition for the main multi-sensor, 16-class configuration of VTT-ConIoT, quantifying the cross-subject performance spread within this benchmark (Section~\ref*{sec:persubject}).
\item A systematic configuration sensitivity map, covering sensor placement, window size, and six classifier families, with a hyperparameter-tuning check (Sections~\ref*{sec:sensor}--\ref*{sec:window}) and a personalised-calibration pilot (Section~\ref*{sec:personalised_calibration}).
\item A validated comparison of \acrshort{gan}-based and standard time-series augmentation, clarifying when each is (and is not) likely to help (Section~\ref*{sec:augmentation}).
\end{enumerate}}

% ============================================================
\section{Related Work}
\label{sec:related}

\subsection{Construction Worker \acrshort{har} on VTT-ConIoT}

The original VTT-ConIoT study \cite{makela2022vtt} reported
\gls{svm} baselines of \SI{78}{\percent}/\SI{89}{\percent} (16-/6-class, \SI{5}{\second} window;
\SI{84}{\percent}/\SI{94}{\percent} at \SI{10}{\second}) under \gls{loso}, and a per-subject breakdown
(\SIrange{39}{62}{\percent}, mean \SI{52}{\percent}) only for its simpler single-sensor 6-class
baseline; a per-subject decomposition for the main multi-sensor,
16-class configuration is absent from the literature.

Five subsequent studies report up to $\approx$\SI{99}{\percent} accuracy on
this benchmark, rarely under genuine subject independence:
\gls{lstm}~\cite{mekruksavanich2022lstm} and
\gls{cnn}-based~\cite{mekruksavanich2023deep} models use $k$-fold
cross-validation without stating subject independence (up to
\SI{98.55}{\percent} and \SI{99.91}{\percent}); \cite{mekruksavanich2025resnet} reports
a participant-grouped 5-fold result (\SI{98.68}{\percent}) vs.\ its own
\emph{strict} \gls{loso} result (\SI{91.20}{\percent}); \cite{lamsa2022video2imu}
applies genuine \gls{loso} only to the simplified
single-sensor 6-class baseline (\SI{52}{\percent}, matching
\cite{makela2022vtt} rather than $\approx$\SI{99}{\percent});
and \cite{chen2025geometry} uses a subject-blind split, reporting
\SIrange{70}{83}{\percent} F1 while citing
\cite{mekruksavanich2023deep}'s \SI{99.9}{\percent} only as related work.
The few studies evaluating subject-independently on a comparable
configuration report figures well below $\approx$\SI{99}{\percent}; none
report per-subject distributions; sensor-placement rankings
conflict across papers; and none examine configuration trade-offs
or augmentation.

Placement optimality is sensitive to protocol, classifier, and
window length~\cite{ubicomp2023placement}, and window size has a
monotonic positive effect across \gls{har} datasets
\cite{springerwindow2025}; \gls{gan}-based
augmentation~\cite{xu2019ctgan} does not reliably outperform
standard time-series transforms~\cite{um2017data,arxiv2506scaling}.
No such sensitivity or augmentation study exists for
construction-specific \gls{har}.

% ============================================================
\section{Dataset and Methods}
\label{sec:methods}

\subsection{VTT-ConIoT Dataset}

VTT-ConIoT~\cite{makela2022vtt} contains data from 13
healthy adult participants (age 25--55; 10\,M/3\,F)
performing 16 construction activities in a controlled
laboratory, grouped into six task families
(painting, cleaning, climbing, hands-up, floor-work,
walking).
At the main \SI{5}{\second}-window configuration the 16 activity classes
are approximately balanced: 715 windows for 15 classes and 660 for the
remaining one (11{,}385 windows total), so no class-imbalance correction
was applied.
Three Aistin iProxoxi sensors were worn simultaneously:
\emph{trouser pocket/hip}, \emph{upper back}, and
\emph{shoulder/hand}, sampled at $\approx$\SI{100}{\hertz}.

\subsection{Preprocessing and Feature Extraction}

Accelerometer, gyroscope, and per-sensor acceleration norm
channels (21 total) were retained; magnetometer and
barometric channels were excluded.
A sliding window approach was applied with a fixed 100-sample
(\SI{1}{\second}) stride across three durations: \SI{2}{\second} (200 samples), \SI{5}{\second}
(500 samples), and \SI{10}{\second} (1000 samples), corresponding to
\SI{50}{\percent}, \SI{80}{\percent}, and \SI{90}{\percent} overlap respectively.
From each window, 147 statistical features were extracted
(mean, median, variance, 25th/75th percentile,
minimum, maximum; 7 statistics $\times$ 7 values $\times$ 3 sensors). Features were standardised within
 each \gls{loso} fold using parameters estimated from the training
 split only, then applied to the test split to prevent data leakage.

\subsection{Classifiers}

Six classifiers were evaluated: \gls{rf},
\gls{et}, \gls{xgb}, \gls{lda}, \gls{lr}, and
\gls{svm} (RBF kernel), all with default scikit-learn
hyperparameters (only the random seed fixed and, where
needed for convergence, the iteration limit increased) to
reflect out-of-the-box deployment. To check whether default
configurations are a fair basis for comparison, we additionally
ran a nested hyperparameter search (\gls{rf}, \gls{xgb},
\gls{svm}, \gls{lr}; inner \acrshort{loso}-consistent grouped
search on training participants only, never touching the held-out
participant) and report the tuned-vs-default delta in
Section~\ref*{sec:sensor}.

\subsection{Evaluation Protocol and Sensor Configurations}

All reported results use a single consistent protocol:
participant-wise \acrshort{loso} cross-validation, 13 folds,
each holding out one participant for testing, with feature
standardisation fit on the training fold only, for every sensor
configuration and window length, avoiding conclusions resting
on the identity of a single held-out subset.

Five sensor subsets were evaluated:
\emph{All} (back\,+\,hip\,+\,hand),
\emph{Back only}, \emph{Hand only},
\emph{Trouser/Hip only}, and
\emph{Trouser accelerometer-only} (ablation).

\subsection{Additional Validation Experiments}
\label{sec:personalization-methods}

\textbf{Personalised calibration}: to test whether target-participant
data closes the per-subject gap (Section~\ref*{sec:persubject}),
each \acrshort{loso} fold was retrained with a calibration slice
of the held-out participant's own data (\SI{0}{\percent}/\SI{10}{\percent}/\SI{20}{\percent}/\SI{30}{\percent}, split by
temporal order per activity to avoid window-overlap leakage)
added to training, evaluated on the remaining disjoint portion.
\textbf{Augmentation}: \gls{ctgan}~\cite{xu2019ctgan} was fit per
activity class, strictly inside the training fold, generating
synthetic samples 1:1 with each class's real training count, on a
6-of-13-fold representative subset (weakest-to-strongest
participants, Section~\ref*{sec:persubject}) at 100 epochs, with
a 10-epoch, 3-seed variance check on two of those folds;
synthetic-vs-real similarity was checked per feature
(Kolmogorov--Smirnov). The
jitter/scaling/time-warp method of Um et al.~\cite{um2017data}
was evaluated identically on this subset and the full 13-fold
protocol, with \gls{rf}, \gls{et}, \gls{lda}, \gls{lr}, \gls{svm}.
\textbf{Deep-learning baseline}: a 1-D \gls{cnn} was trained on
raw \SI{5}{\second} windows (21 channels $\times$ 500 timesteps) per fold,
with class-weighted loss, dropout, batch normalisation, and early
stopping on a training-participant-only validation split.

% ============================================================
\section{Results}
\label{sec:results}

\subsection{Per-Subject \acrshort{loso} Decomposition}
\label{sec:persubject}

\begin{figure}[t]
  \centering
  \includegraphics[width=\columnwidth]{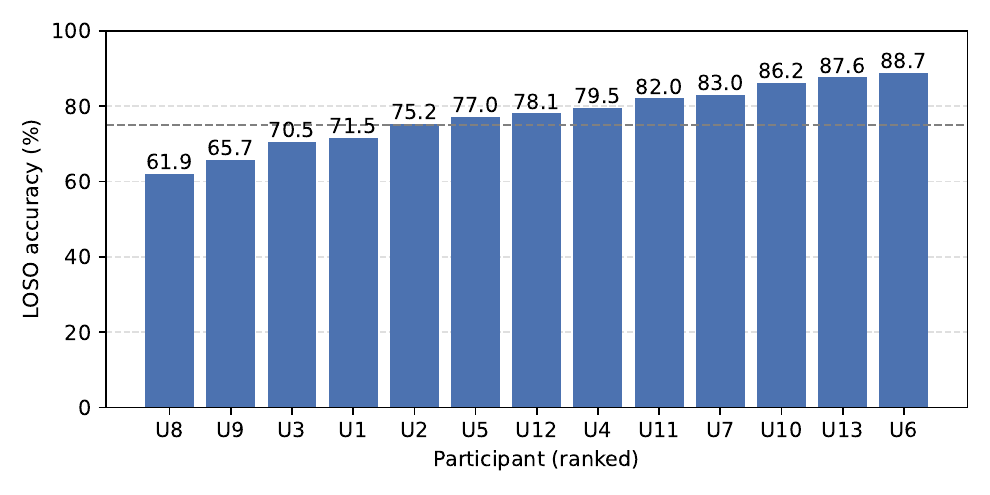}
  \caption{Per-subject \acrshort{loso} accuracy, ranked lowest to
           highest, main configuration (\MainClassifier{}, all sensors,
           \SI{5}{\second}); dashed line marks an illustrative \SI{75}{\percent}
           grouping aid, not a deployment threshold.}
  \label{fig:persubject}
\end{figure}

Fig.~\ref*{fig:persubject} shows per-participant \gls{loso}
accuracy for the all-sensors, \SI{5}{\second} window, \MainClassifier{}
configuration.
Once every classifier is correctly feature-scaled per fold,
\MainClassifier{} is the best-performing classifier for this
configuration.
Individual accuracies range from \textbf{\SIrange{\PerSubjMin}{\PerSubjMax}{\percent}}
(mean $\SI{\PerSubjMean}{\percent} \pm \SI{\PerSubjStd}{\percent}$,
macro-F1 \SI{\MainMacroFScore}{\percent}), a \SI{\PerSubjSpread}{\pp}
spread. This aggregate is our own \gls{loso} result for the main
multi-sensor, 16-class configuration, well below the
$\approx$\SI{99}{\percent} headline figures often quoted for this benchmark
(Section~\ref*{sec:related}).

This spread shows that a single aggregate statistic, even our
own genuinely subject-independent one, conceals large
per-subject variability: the weakest quartile (Users
\WeakUserList{}) sits at \SIrange{\WeakAccMin}{\WeakAccMax}{\percent}, the
strongest (Users \StrongUserList{}) at
\SIrange{\StrongAccMin}{\StrongAccMax}{\percent}. The illustrative \SI{75}{\percent} line
in Fig.~\ref*{fig:persubject} aids visual grouping only, with
no clinical, regulatory, or operational basis. For field
deployment, the worst-case individual, not the mean, is the
operationally relevant figure: \SI{\PerSubjMin}{\percent} is well above the
\SI{6.25}{\percent} random baseline for 16 classes, but far enough below the
cohort mean that this worker would be systematically
under-served by a one-size-fits-all model.

\subsection{Sensor Configuration and Classifier Comparison}
\label{sec:sensor}

Table~\ref*{tab:sensor} reports \gls{loso} mean accuracy across
five sensor placements and six classifiers at the \SI{5}{\second} window,
under the same protocol throughout.
Multi-sensor fusion is strongest for every classifier: for
\MainClassifier{}, fusing all sensors gains
\SI{\AllVsBestSingleGainPP}{\pp} over the best single sensor
(\BestSingleSensor{}); back and hand perform comparably as
single sensors, both clearly outperforming trouser. Restricting
the trouser sensor to its 4 accelerometer channels (dropping
gyroscope) reduces accuracy by \SI{\TrouserAccDropPP}{\pp} for
\MainClassifier{}. Pooling the main configuration's confusion matrix
across all 13 folds, the two most confused task-family pairs are
\TopConfusedPairA{} and \TopConfusedPairB{}. These activities
share overlapping whole-body motion signatures.

\textbf{Hyperparameter Tuning}: Separately, a nested hyperparameter
search found a tuned-vs-default delta of at most
\SI{\MaxHparamDelta}{\pp} (and for \gls{rf}, tuning did not outperform
defaults at all), indicating defaults are not a materially unfair
basis for this classifier comparison.

\textbf{Deep-Learning Baseline}: A class-weighted, regularised 1-D
\gls{cnn} achieved only \SI{\CnnMeanAcc}{\percent} mean accuracy
(macro-F1 \SI{\CnnMeanMacroFScore}{\percent}) under the same protocol,
clearly below the classical classifiers. Classical feature-based
classifiers thus remain more sample-efficient than a from-scratch
\gls{cnn} here, far below the $\approx$\SI{99}{\percent}
deep-learning figures reported elsewhere for this dataset,
underscoring that those figures are not obtained under genuine
subject-independent evaluation (Section~\ref*{sec:related}).

\begin{table}[t]
\caption{LOSO Mean Accuracy (\si{\percent}) over 13 folds, by Sensor Configuration and Classifier (\SI{5}{\second} window; std \SIrange{6}{11}{\pp})}
\label{tab:sensor}
\centering\scriptsize
\begin{tabular}{lccccc}
\toprule
\textbf{Clf.} & \textbf{All} & \textbf{Back} & \textbf{Hand} & \textbf{Trouser} & \textbf{Trsr-Accel} \\
\midrule
\acrshort{svm} & \textbf{77.5} & \textbf{67.9} & 66.4 & 53.1 & 41.8 \\
\acrshort{et} & 76.6 & 66.8 & 64.9 & 52.1 & 37.4 \\
\acrshort{rf} & 76.4 & 67.1 & 65.1 & \textbf{55.0} & 38.2 \\
\acrshort{xgb} & 75.9 & 67.0 & \textbf{66.6} & 54.3 & 38.4 \\
\acrshort{lda} & 73.3 & 59.8 & 60.2 & 49.4 & 42.7 \\
\acrshort{lr} & 70.8 & 63.3 & 62.8 & 53.3 & \textbf{46.9} \\
\bottomrule
\end{tabular}
\end{table}

\subsection{Window Size Sensitivity}
\label{sec:window}

\begin{figure}[t]
  \centering
  \includegraphics[width=\columnwidth]{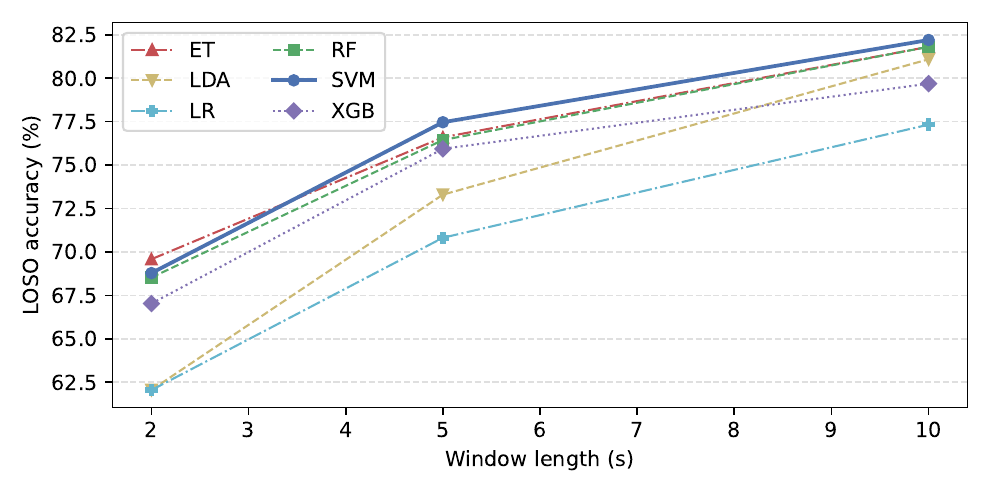}
  \caption{\acrshort{loso} accuracy vs.\ window length (all sensors), one
           line per classifier; \MainClassifier{} (bold) is the
           main-text classifier.}
  \label{fig:window}
\end{figure}

Fig.~\ref*{fig:window} shows accuracy as a function of
window length for all six classifiers using all sensors, under
the same \gls{loso} protocol as Table~\ref*{tab:sensor}.
Every classifier improves monotonically with window length; for
\MainClassifier{}, accuracy rises from \SI{\WinTwoAcc}{\percent} (\SI{2}{\second}) to
\SI{\WinFiveAcc}{\percent} (\SI{5}{\second}) to \SI{\WinTenAcc}{\percent} (\SI{10}{\second}), a
\SI{\WinGainPP}{\pp} gain from \SI{2}{\second} to \SI{10}{\second}.
This trend is consistent with the general \acrshort{har}
literature~\cite{springerwindow2025}: longer windows capture
complete activity cycles (e.g., a full paint stroke, stair
ascent) and produce more stable statistical estimates.
For real-time ergonomic alert systems the \SI{5}{\second} window offers a
strong balance between accuracy and feedback latency; \SI{2}{\second}
remains the only window among those tested that comes close when
the fastest available response is required, at a real accuracy cost.

\subsection{Synthetic Data Augmentation}
\label{sec:augmentation}

Table~\ref*{tab:ctgan}
reports the \gls{ctgan}-augmented comparison: \gls{ctgan} was fit
per activity class, strictly inside each held-out participant's
training fold, on a 6-of-13-fold representative subset (the four
weakest, one median, one strongest participant per
Section~\ref*{sec:persubject}), at 100 epochs; standard
jitter/scaling/time-warp augmentation~\cite{um2017data} was
evaluated on the same subset and, as a corroborating check, on
the full 13-fold protocol (\SIrange{\StdAugDeltaMinFull}{\StdAugDeltaMaxFull}{\pp},
consistent in sign and magnitude).

\begin{table}[t]
\caption{Augmentation Delta (6-of-13-fold LOSO subset, users 8,5,1,9,2,6, \SI{5}{\second} window)}
\label{tab:ctgan}
\centering\scriptsize
\begin{tabular}{lccccc}
\toprule
Clf. & Base & +StdAug & $\Delta$ & +CTGAN & $\Delta$ \\
\midrule
RF & 73.3 & 73.5 & +0.2 & 72.1 & -2.0 \\
ET & 73.0 & 73.6 & +0.6 & 73.6 & -0.1 \\
LDA & 69.7 & 69.5 & -0.1 & 63.6 & -5.5 \\
LR & 66.1 & 66.8 & +0.7 & 64.5 & -1.4 \\
SVM & 73.3 & 74.1 & +0.7 & 72.4 & -1.4 \\
\bottomrule
\end{tabular}
\end{table}

On the shared 6-fold subset (Table~\ref*{tab:ctgan}), the two
methods diverge sharply: standard augmentation gives small,
mostly positive gains (\SIrange{\StdAugDeltaMinSub}{\StdAugDeltaMaxSub}{\pp}),
while \gls{ctgan} \emph{degrades} accuracy for every classifier
(\SI{\CtganDeltaMax}{\pp} to \SI{\CtganDeltaMin}{\pp}, worst for
\CtganWorstClf{}). A per-feature Kolmogorov--Smirnov test found
\CtganKSSigCount{}/\CtganKSTotalCount{} (\SI{\CtganKSSigPct}{\percent})
feature comparisons distinguishable from real data at $p<0.05$:
the synthetic samples are not distributionally faithful,
consistent with \gls{ctgan} treating each time-step as an
independent tabular row and losing \gls{imu} windows'
autocorrelation. A 3-seed check on the weakest and strongest
participants varied by only a few points, so this is not a
single unlucky run. This sharpens prior evidence that generative
augmentation does not reliably improve on simpler, physically
motivated approaches for \acrshort{imu}-based
\acrshort{har}~\cite{arxiv2506scaling}. Here, it is actively
counter-productive.

\subsection{Personalised Calibration}
\label{sec:personalised_calibration}
The per-subject gap is concentrated in a minority of participants rather than distributed uniformly across the cohort. We therefore tested whether a small amount of target-participant data could narrow this gap. Among the four weakest participants, mean accuracy increased from \SI{\PersWeakAccNone}{\percent} without calibration data to \SI{\PersWeakAccThirty}{\percent} after adding \SI{30}{\percent} of each participant's own temporally disjoint data to the training set, recovering \SI{\PersWeakGainPP}{\pp}. This result identifies low-data personalised calibration as a practical mitigation for one-size-fits-all deployment.

% ============================================================
\section{Discussion}
\label{sec:discussion}

\subsection{Cross-Subject Variability and Deployment Guidance}

Our results show that the mean accuracy alone conceals substantial
variation across participants. The worst-case participant
represents workers whose movement, morphology, or task style
deviate from the training distribution. This is precisely the
population for whom monitoring is most safety-critical, and whom
a mean-only report would hide. Overlapping-window leakage is a
distinct, compounding pitfall reported across the \gls{har}
field~\cite{pmc2025shift}, which is why every result here uses
grouped \gls{loso} rather than a hold-out split. For
practitioners: report per-subject distributions, not aggregates;
prefer multi-sensor fusion, which benefits more here than in
feature-reduced configurations \cite{mekruksavanich2025resnet}; and
favour a \SIrange{5}{10}{\second} window.

\subsection{Augmentation, Personalisation, and Model Choice}
The augmentation results favour simple, physically motivated transformations over \gls{ctgan}-style synthesis. Jitter, scaling, and time-warping~\cite{um2017data} offer a low-cost first step for expanding this \gls{imu} feature representation. In contrast, \gls{ctgan} actively degrades accuracy, with the distributional-fidelity analysis providing a likely explanation. \gls{gan}-based approaches may still be appropriate for the separate, unvalidated goal of synthesising new \emph{movement styles}, rather than additional samples from existing participants.
\begin{comment}
Apply standard time-series augmentation (jitter, scaling,
time-warp~\cite{um2017data}) as a cheap, physically motivated
first step. Do not default to \gls{ctgan}-style feature-level
augmentation for this kind of \gls{imu} representation: it
actively degrades accuracy, with a demonstrated
distributional-fidelity mechanism behind the harm.
\gls{gan}-based approaches may still suit a different,
unvalidated goal: synthesising new \emph{movement styles} absent
from the training population, not additional samples from
existing participants.
\end{comment}

Personalised calibration narrows the per-subject gap for the weakest participants with minimal data (Section~\ref*{sec:personalised_calibration}), and hyperparameter tuning yielded at most \SI{\MaxHparamDelta}{\pp} over defaults (Section~\ref*{sec:sensor}), confirming the classifier comparison is fair.
% ============================================================
\section{Conclusion}
\label{sec:conclusion}

We present, to our knowledge, the first systematic configuration
sensitivity analysis of the VTT-ConIoT construction worker
\acrshort{imu} benchmark under a single consistent \acrshort{loso}
protocol. Our per-subject decomposition reveals a
\SI{\PerSubjSpread}{\pp} cross-subject spread
(\SIrange{\PerSubjMin}{\PerSubjMax}{\percent}, mean \SI{\PerSubjMean}{\percent}), concealed
by the $\approx$\SI{99}{\percent} headline figures quoted elsewhere for
this benchmark. Multi-sensor fusion, a \SIrange{5}{10}{\second} window,
and properly-scaled classical classifiers form a robust baseline;
standard augmentation gives small, consistent gains while
\gls{ctgan} augmentation harms accuracy; and personalised
calibration recovers \SI{\PersWeakGainPP}{\pp} for the weakest
participants from just \SI{30}{\percent} of their own data, and a
from-scratch \gls{cnn} baseline remains behind the classical
ensembles under this same protocol.

% ============================================================
\section*{Acknowledgement and Code Availability}

The authors thank VTT Technical Research Centre of Finland
for making the VTT-ConIoT dataset publicly available.
Code and aggregated results to reproduce the experiments in this
paper are available at
\url{https://github.com/IhsanUlHaq/bsn2026-imu-har}.

% ============================================================
% ponytail: page-budget nudge, not a content change -- restoring the CNN
% baseline paragraphs (Methods + Results) pushed p.4's last column just
% past flush-full again, with only ref. [13] (~3 lines) spilling to p.5;
% same safe-trim-first approach as before (see git log) closed most of
% it, this covers the rest. Re-check/adjust if the surrounding text
% changes.
\enlargethispage{40pt}
{\scriptsize
\bibliographystyle{IEEEtran}
\bibliography{references_bsn}
}

\end{document}